\documentclass[journal]{IEEEtran}
\makeatletter
\DeclareRobustCommand\onedot{\futurelet\@let@token\@onedot}
\def\@onedot{\ifx\@let@token.\else.\null\fi\xspace}

\def\etc{\emph{etc}\onedot} 
 
\def\etal{\emph{et al}\onedot}
\makeatother
\usepackage[boxed,ruled,commentsnumbered]{algorithm2e}
\usepackage{amssymb}
\usepackage{mathtools}
\ifCLASSINFOpdf
\else
\fi
\begin{document}
%
\title{Attend to the Difference: Cross-Modality Person Re-identification via Contrastive Correlation}
%
%
%

\author{\IEEEauthorblockN{Shizhou Zhang{*},
Yifei Yang{*},
Peng Wang,
Guoqiang Liang\dag,
Xiuwei Zhang, and \\
Yanning Zhang,~\IEEEmembership{Senior Member,~IEEE}}

\thanks{
Shizhou Zhang, Yifei Yang, Peng Wang, Guoqiang Liang, Xiuwei Zhang, and Yanning Zhang are with National Engineering Laboratory for Integrated Aero-Space-Ground-Ocean Big Data Application Technology, School of Computer Science and Engineering, Northwestern Polytechnical University, Xi'an 710071, China.
This work is partially supported by the National Natural Science Foundation of China (No.62101453, No.61902321, No.U19B2037) and the National Key R\&D Program of China (No. 2020AAA0106900).
}
\thanks{
* indicates co-first authors. \dag
Corresponding author: Guoqiang Liang (gqliang@nwpu.edu.cn).}}

%
%

\markboth{IEEE Transactions on Image Processing}%
{Shell \MakeLowercase{\textit{et al.}}: Bare Demo of IEEEtran.cls for IEEE Journals}
%



\maketitle

\begin{abstract}
The problem of cross-modality person re-identification has
been receiving increasing attention recently, due to its practical significance.
Motivated by the fact that human usually attend to the difference when they compare two similar objects,
we propose a dual-path cross-modality feature learning framework which preserves intrinsic spatial structures and attends to
the difference of input cross-modality image pairs.
Our framework is composed by two main components: a Dual-path Spatial-structure-preserving Common
Space Network (DSCSN) and a Contrastive Correlation Network (CCN).
The former embeds cross-modality images into a common 3D tensor space without losing spatial structures,
while the latter extracts contrastive features by dynamically comparing input image pairs.
Note that the representations generated for the input RGB and Infrared images are mutually dependant to each other.
We conduct extensive experiments on two public available RGB-IR ReID
datasets, SYSU-MM01 and RegDB, and our proposed method
outperforms state-of-the-art algorithms by a large margin with
both full and simplified evaluation modes.
\end{abstract}

\begin{IEEEkeywords}
Cross-modality, Re-identification, Common Space, Contrastive Correlation.
\end{IEEEkeywords}

%
\IEEEpeerreviewmaketitle

\section{Introduction}
%
%
%
%
\IEEEPARstart{P}{erson} Re-IDentification (ReID) in RGB cameras has been attracting wide attentions in recent years as it is of significance in video surveillance~\cite{zheng2016person,li2018adversarial,wei2018person,zheng2019joint,hou2019vrstc,yu2019unsupervised,zhao2019attribute,li2018toward,yang2018person,zhou2019person,xu2015distance}.
However, these algorithms are all developed with an assumption of good lighting conditions.
Nowadays, many surveillance cameras support automatic switching between RGB and infrared working modes according to the surrounding illumination~\cite{wu2017rgb}.
Such real-world scenarios urge the research requirement of RGB-Infrared cross modality person ReID (RGB-IR ReID), where given an RGB (infrared) image of a query person, it aims to match the infrared (RGB) images of the same person in other disjoint cameras~\cite{wu2017rgb,ye2018hierarchical,wang2019learning,dai2018cross,ye2018visible}.

The RGB-IR ReID task suffers from the difficulties of both cross modality discrepancy resulting from the imaging sensors and intra-modality appearance discrepancy due to illumination, background, pose and viewpoint variations, just as in the traditional ReID task.
As stated in~\cite{wang2019learning}, the cross modality discrepancy is often more challenging than the intra-modality appearance discrepancy.
To alleviate the cross modality discrepancy, a popular pipeline which includes feature extraction phase and feature embedding phase is introduced. For feature extraction phase, a multi-branch architecture is adopted to extract modality specific feature vectors firstly, and for feature embedding phase, a mapping function is then adopted to project the modality specific features into a common feature space~\cite{ye2018visible,wang2017vqa}.
Generally speaking, the model branches used to extract modality specific features are not required to have same architectures or share parameters necessarily, as optimal feature extraction model highly depends on the input data modalities.
Note that in the feature extraction phase, the modality specific feature is usually processed into a 1D-shaped vector, and then the mapping function is typically devised as one or several fully-connected layers to project the modality specific feature vectors, leading to a common feature space which loses the spatial structure information as it is spanned by 1D-shaped feature vectors.
\begin{figure*}[ht!]
	\centering
		\includegraphics[width=1.0\linewidth]{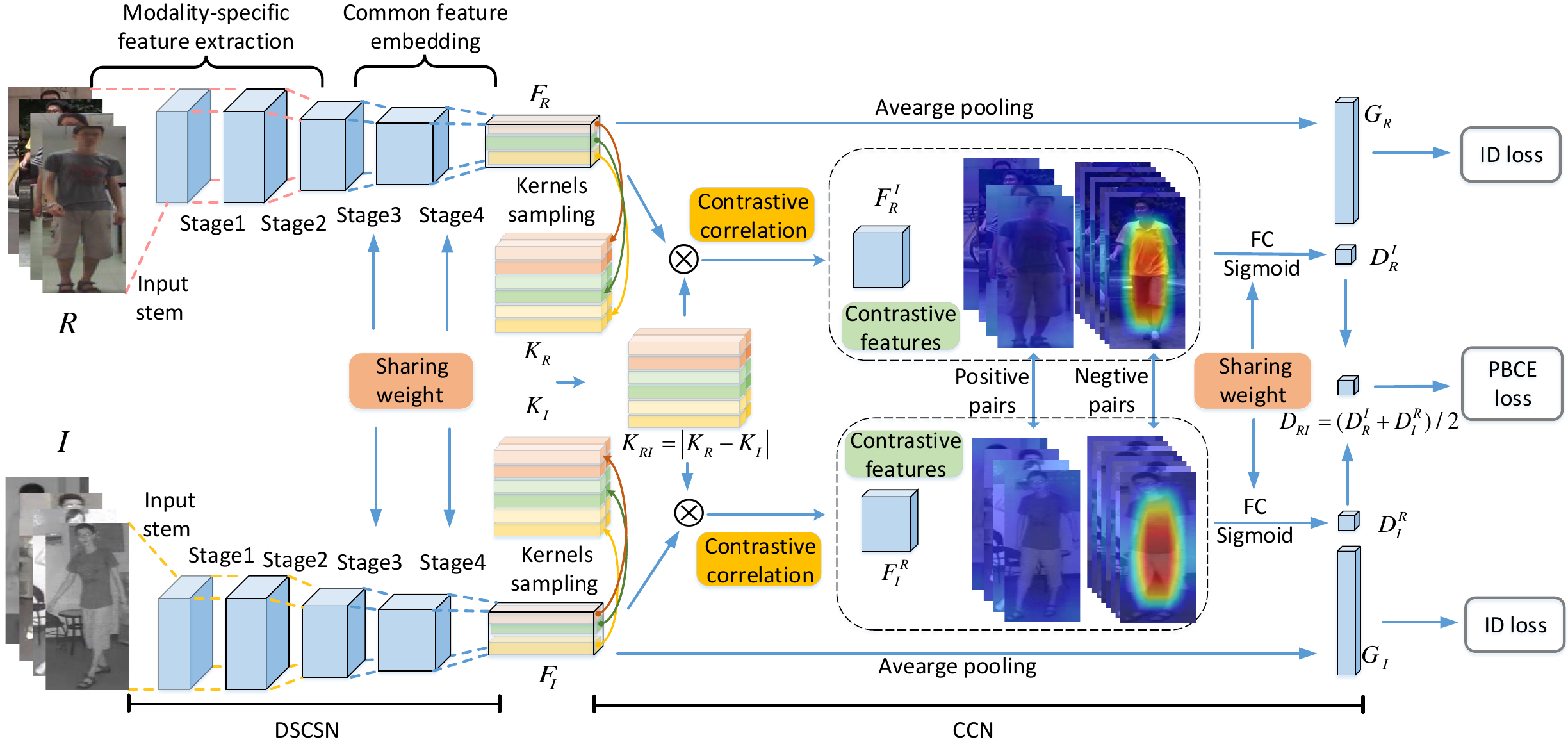}
	\caption{The pipeline of our proposed framework for RGB-IR person ReID. It mainly contains two components: a Dual-path Spatial-structure-preserving Common Space Network (DSCSN) and a Contrastive Correlation Network (CCN).}
	\label{fig:arch}
\end{figure*}

As we all know,
when humans compare two input images, the observation of one input is guided by that of the other, i.e. finding the differences and putting more attention on them for better distinguishing of the two input person images.
However, a common feature space spanned by 1D-shaped feature vectors does not contain the spatial structures anymore.
In this paper, motivated by the fact stated above,
we propose a dual path cross modality feature learning framework which consists of a Dual-path Spatial-structure-preserving Common Space Network (DSCSN) and a Contrastive Correlation Network (CCN).
For the DSCSN, the modality specific feature space and the common feature space are devised to be 3D-shaped tensor spaces which can still preserve the spatial structures as supported by convolution feature maps.
Furthermore, benefiting from the spatial-structure-preserving common feature space, the CCN is incorporated to make the final feature representations attend to the difference of the input RGB-IR person image pairs.
Specifically, the kernels of CCN are generated by a parameter-free kernel generator module.
Firstly, personalized kernels of each input person are sampled from the common feature representations.
Then the difference of the personalized kernels is exploited to serve as the contrastive correlation kernels, which are expected to attend to the difference between the input person pairs.
As can be seen from Figure~\ref{fig:arch}, the proposed method can be viewed as a pipeline that the input RGB (Infrared) image can obtain dynamic feature representations that depend on which Infrared (RGB) image it is compared with~\cite{han2018face}.
During inference, the similarities between two input images can be calculated based on a full mode or a simplified mode.
Extensive experiments are conducted on two public available RGB-IR ReID dataset, SYSU-MM01~\cite{wu2017rgb} and RegDB~\cite{nguyen2017person} and our proposed method achieves state-of-the-art performances for both the full and simplified evaluation modes.

To summarize, the main contributions of this paper include:
\begin{itemize}
    \item We design a Dual-path Spatial-structure-preserving Common Space Network (DSCSN) to embed cross-modality images into a common feature space. In order to preserve the intrinsic spatial structure, the common feature space (as well as intermediate modality specific feature space) are represented by 3D tensors rather than conventionally adopted 1D vectors. Experiments show that DSCSN is superior to traditional dual-path architectures for RGB-IR ReID.
    \item To capture the semantic difference between input RGB-IR image pairs, we design a Contrastive Correlation Network (CCN), which dynamically builds contrastive kernels in a parameter-free fashion and then apply them over the entire image feature map by cross-correlation. The resulting contrastive features present high responses on regions with clear semantic differences.
    \item Extensive experiments on the popular SYSU-MM01 and RegDB datasets demonstrate that our proposed approach outperforms existing algorithms by a large margin.
\end{itemize}

\section{Related Work}\label{sec_RW}
In this section, we review the literature related to our work from the following two aspects.

\noindent\textbf{Cross Modality ReID.} For cross modality person ReID, some endeavors are devoted on text-to-image person retrieval~\cite{li2017person,zheng2017dual,zhou2017attention}, whose approaches can not be directly transferred to RGB-IR ReID problem  due to the obvious difference between the two tasks.
More related to our work, some pioneer researches have explored the problem of RGB-IR ReID.
Wu \etal{}~\cite{wu2017rgb} provided the first RGB-IR cross modality ReID dataset named SYSU-MM01 for the community and proposed a deep zero-padding framework for automatically evolving domain-specific structure for RGB-IR matching.
Ye \etal{}~\cite{ye2018hierarchical} put forward a hierarchical learning framework to jointly optimize the modality-specific and modality-shared metrics.
Ye \etal{}~\cite{ye2018visible} further introduced a dual-path network trained with a bidirectional dual-constrained top-ranking loss to handle both the cross modality and intra-modality variations.
Learning discriminative modality-invariant feature representations in an adversarial way is proposed by Dai \etal{}~\cite{dai2018cross}.
Wang \etal{}~\cite{wang2019learning} presented a dual-level discrepancy reduction learning scheme which incorporated a generative model to transform the input into a unified space firstly, then extracted discriminative features from the unified space.
In our method, we also adopt a dual-path CNN similar as in~\cite{ye2018hierarchical,ye2018visible,dai2018cross}, while in our architecture the modality specific feature space and the common feature space are devised to be 3D tensor spaces which can still preserve the spatial structures.
Benefiting from this, a Contrastive Correlation Network can be introduced to make the final features attend to the difference of the input RGB-IR person image pairs.

\noindent\textbf{Dynamic Convolution Kernels.}
Our proposed Contrastive Correlation Network is related to adaptive convolution operations with dynamic kernels.
HyperNetworks~\cite{ha2016hypernetworks} incorporated a hypernetwork to generate weights for a main network and the hypernetwork usually takes information about the structure of the weights as inputs.
~\cite{kang2017incorporating} proposed to take advantage of side information such as camera angle and height to generate the convolution kernels adaptive to the context.
~\cite{jia2016dynamic} proposed the dynamic filter network for the task of video and stereo prediction, where the filters are dynamically generated conditioned on the input frame.
~\cite{li2018high} formulated the visual tracking task as a one-shot detection problem and proposed the Siamese-RPN tracker where the template branch predicts the ``correlation'' weights for region proposal subnetwork on detection branch.
Most neural models with dynamic convolution kernels take one input to generate the weights, while~\cite{han2018face} proposed to generate contrastive kernels based on a pair of inputs to characterize the specific feature of one input compared with another.
With respect to~\cite{han2018face}, the main differences lie in three folds.
1) The kernel generator module. Our proposed Contrastive Correlation Network utilizes a parameter-free sampling scheme to replace the FC layer of the kernel generator module in~\cite{han2018face} which greatly reduces the number of parameters relative to the whole model and the experimental results showed that the performance will not be hampered and our method can more easily get converged; 2) The inference settings. During inference phase, we propose both full and simplified evaluation modes to calculate the similarities  between two input images; 3) The modalities and views of input data. ~\cite{han2018face} aims to verify the input face image pairs in single modality and the view changes are relatively tempered, while our method is conducted on cross modality person matching and the view changes are relatively drastic as they are from different cameras.

\section{Proposed Method}\label{sec_Method}

In this section, we elaborate the framework of the proposed feature learning method for RGB-IR person Re-Identification.
The framework learns more discriminative feature representations by projecting two modalities into a spatial-structure-preserving common space and attending to the differences between the paired feature maps.

As shown in Figure~\ref{fig:arch}, it comprises two main components: a Dual-path Spatial-structure-preserving Common Space Network (DSCSN) and a Contrastive Correlation Network (CCN).
Specifically, the DSCSN utilizes a partially shared structure to learn the RGB-IR spatial-structure-preserving common features by simultaneously modeling the modality specific and modality shared information.
The Contrastive Correlation Network is to mimic the mechanism of learning by attending to the difference of two similar objects.
\subsection{Dual-path Spatial-structure-preserving Common Space Network}

A Dual-path Spatial-structure-preserving Common Space Network (DSCSN) is designed to extract common features with the shape of 3D convolution feature maps for the input RGB and IR images.
It consists of two branches: RGB-branch and IR-branch and both the branches are designed to have similar network structures.
Note that it mainly contains two steps: modality specific feature extraction and common feature embedding. Feature extraction step focuses on capturing modality specific information for different image modalities, and the feature embedding step aims to learn common features of RGB and IR image modalities.

As shown in Figure~\ref{fig:arch}, RGB images and IR images are fed into the DSCSN. The low-level layers without sharing parameters are designed as the feature extraction part to extract the modality-specific features.
After that, convolution architectures with shared parameters on top of feature extraction part are treated as a common feature embedding function to project the modality-specific inputs into a common space spanned by 3D convolution feature maps. To clarify,
$C_{R}(\cdot)$ and $C_{I}(\cdot)$ are denoted as transformation functions from the input images to common space features
for RGB images and IR images respectively.
Given an RGB image $R$ and an IR image $I$,
the extracted 3D common features $F_{R}$ and $F_{I}$ are represented by
\begin{equation}
\begin{split}
    F_{R}=C_{R}(R)\in \mathbb{R}^{h_{F}\times w_{F}\times c_{F}} \\
    F_{I}=C_{I}(I)\in \mathbb{R}^{h_{F}\times w_{F} \times c_{F}}
\end{split}
\end{equation}
where $h_{F}$, $w_{F}$ and $c_{F}$ are the height, width, and number of channels respectively.

It is worth mentioning that different from dual-path network proposed in~\cite{ye2018visible}, which introduces a shared FC layer as feature embedding function acting on 1D feature vectors, our feature embedding adopts convolutional architectures and acts on 3D feature tensors. The proposed network can preserve some spatial structure information for the common space.
From Figure~\ref{fig:arch} it can be seen that although RGB and IR images undergo different imaging process, they still have some similar spatial visual patterns.
Experimental results in Section~\ref{sec_Exp} show that our feature tensor embedding method achieves better performance than~\cite{ye2018visible} for RGB-IR ReID.

Furthermore, benefiting from the spatial-structure-preserving common space, the Contrastive Correlation Network is further incorporated to make the feature learning model attend to the difference of the input pairs, which will be elaborated in the next subsection.

\subsection{Contrastive Correlation Network}
After obtaining the common feature tensors extracted by the DSCSN, the Contrastive Correlation Network is introduced to force the model to attend to the difference between the paired person images.
The Contrastive Correlation Network consists of two modules: kernel generator module and contrastive correlation module. Here and after, the kernel in Contrastive Correlation Network denotes the correlation filter.

The kernel generator module produces personalized kernels for a person image, aiming at emphasizing the distinct features of a person relative to the mean person. Then the contrastive kernels are calculated as the difference between two personalized kernels.
The contrastive correlation module extracts dynamic contrastive feature of a person image based on which image it is compared with, by performing correlation with the contrastive kernels. Note that the conventional CNN-based methods~\cite{ye2018visible,dai2018cross,wang2019learning} use the same feature of a person image no matter which image it is compared with, once the training of the model is finished.

\noindent\textbf{Kernel generator. }
The kernel generator aims at producing kernels specific to $I$ or $R$, which is referred to as personalized kernels.

Taking the RGB image $R$ as an example,
each kernel tries to illustrate the characteristic of a local part of person image $R$. A kernel can be obtained by sampling/cropping $F_{R}$ at location $(i, j)$,
\begin{equation}
    K_{R}^{ij}=\mathrm{Cropping}(F_{R},i,j,h_{K},w_{K}),
\end{equation}
where $K_{R}^{ij}$ is a local patch cropped from $F_{R}$ with the height of $h_{K}$, and width of $w_{K}$, and $(i,j)$ are the coordinates of $F_{R}$ indicating where the kernel is cropped.
$\mathrm{Cropping(\cdot)}$ denotes the cropping operation.
$\mathcal{K}_{R}$ denotes a set of kernels densely sampled from $F_{R}$ with a sliding stride,
\begin{equation}
    \mathcal{K}_{R}=\{K_{R}^{11},\cdots,K_{R}^{1M},\cdots,K_{R}^{ij},\cdots,K_{R}^{MN}\}
\end{equation}
The kernel sampling algorithm 
is detailed in Algorithm~\ref{alg:sampling}, where $stride\_v$ and $stride\_h$ denote the vertical and horizontal strides.

\begin{algorithm}[t]
\SetKwInOut{Input}{\textbf{Input}}
\SetKwInOut{Output}{\textbf{Output}}
\Input{$F_{R}, h_{K}, w_{K}, stride\_h, stride\_v$  ;}
\Output{$\mathcal{K}_{R}$; }

$\mathcal{K}_{R} = \{\}$\;
\For { $i=0$; $i+h_{K}<=h_{F}$; $i = i+stride\_v$}
{
    \For {$j=0$; $j+w_{K}<=w_{F}$; $j=j+stride\_h$}
    {
        $K_{R}^{ij}=\mathrm{Cropping}(F_{R},i,j,h_{K},w_{K})$\;
        $\mathcal{K}_{R}+=\{K_{R}^{ij}\}$\;
    }
}
\caption{Kernel Sampling Algorithm}
\label{alg:sampling}
\end{algorithm}

Similarly, we can get the personalized kernels of the input IR image $\mathcal{K}_{I}$.
The personalized kernels sampled from the 3D common feature tensors are expected to capture the intrinsic features of a person, regardless of illuminations, poses, view angles, modalities \etc.

To compute the difference of personalized kernels of two person images, the contrastive kernels are defined as:
\begin{equation}\label{equ:T}
    \mathcal{K}_{RI}=|\mathcal{K}_{R}-\mathcal{K}_{I}|
\end{equation}
The contrastive kernels of the RGB (IR) image are dynamically generated depending on the paired IR (RGB) image, expecting to attend to their difference between the input paired persons, while kernels used in conventional correlation are generated by separate input images.

\noindent\textbf{Contrastive correlation.}
The contrastive features between $R$ and $I$ are computed by taking the correlation between $F_{R}$ and $F_{I}$ with contrastive kernels $\mathcal{K}_{RI}$ respectively as:
\begin{equation}
\label{eqn_ccn}
\begin{split}
 F_{R}^{I} = \mathcal{K}_{RI} \bigotimes F_{R}\\
 F_{I}^{R} = \mathcal{K}_{RI} \bigotimes F_{I}
\end{split}
\end{equation}
where $F_{R}^{I}$ represents the contrastive features of $R$ contrasted with $I$, and $F_{I}^{R}$ is the contrastive features of $I$ contrasted with $R$.
Meanwhile, $\bigotimes$ denotes the correlation operation.


After obtaining the contrastive features of $R$ and $I$, a FC layer followed by sigmoid activation function is deployed to calculate the difference score $D_{R}^{I}$ and $D_{I}^{R}$ between $R$ and $I$ as:
\begin{equation}\label{equ:diff_IV}
\begin{split}
 D_{R}^{I} = \sigma(W_{D} \cdot F_{R}^{I}) \\
 D_{I}^{R} = \sigma(W_{D} \cdot F_{I}^{R})
 \end{split}
\end{equation}
where $\sigma (\cdot)$ is the sigmoid function with $\sigma (x) = \frac{1}{1+e^{-x}}$, and $W_{D}$ denotes the weights of the FC layer.
The overall difference score $D_{RI}$ between $R$ and $I$ is defined as the average of the two difference scores,
\begin{equation}\label{equ::diff_total}
 D_{RI} = (D_{R}^{I}+D_{I}^{R})/2
\end{equation}
The higher $D_{RI}$, the greater the difference between the paired images, and the lower the probability that these images belong to the same person.

\subsection{Overall Loss Function}
Two types of losses are enforced on the proposed model, namely Pairwise BCE loss and ID loss.

\noindent \textbf{Pairwise BCE loss.}
Note that the difference score $D_{RI}$ of the same person is expected to be $0$, while for the different persons it is expected to be $1$. To minimize the difference score of same person pairs and maximize the difference score of different person pairs, a Pairwise Binary Cross Entropy (PBCE) loss is adopted as:
\begin{equation}
 \mathcal{L}_{PBCE} = - \frac{1}{M} \sum_{I,R} [l_{RI}\log(D_{RI})+(1-l_{RI})\log(1-D_{RI})]
\end{equation}
where $l_{RI}$ is the label for the input RGB-IR person pair, with $l_{RI}=0$ representing that $I$ and $R$ are the same person, and $l_{IR}=1$ denoting that $I$ and $R$ are different persons, and $M$ representing the number of person pairs.

\noindent\textbf{ID loss. }
Meanwhile, after performing a global average pooling layer on $F_{R}$ and $F_{I}$, global features $G_{I}$ and $G_{R}$ are obtained for $I$ and $R$ respectively.
For each person, its own characteristics indicates that the features of same persons should have high similarity even if with various pose, illumination, view angle changes and \etc.
So we enforce the identification (ID) loss on top of the global features as in the following equations:
\begin{equation}\label{equ::soft_R}
 p_{R} = \mathrm{softmax}(W_{ID} \cdot G_{R})
\end{equation}
\begin{equation}\label{equ::soft_I}
 p_{I} = \mathrm{softmax}(W_{ID} \cdot G_{I})
\end{equation}
\begin{equation}\label{equ::diff_total}
\begin{split}
 \mathcal{L}_{ID} =   -\frac{1}{2N}\left[\sum_{R}\sum_{c=1}^{C}y^{c}_{R} \log p^{c}_{R}+ \sum_{I}\sum_{c=1}^{C}y^{c}_{I}\log p^{c}_{I}\right]
\end{split}
\end{equation}
where $W_{ID}$ is the weights of the last FC layer of ID loss. $p_{R}$ and $p_{I}$ are the predicted label probability distributions of $R$ and $I$. C is the number of person identities. $y_{R}$ and $y_{I}$ are one-hot coding ID labels for $R$ and $I$ respectively, and N is the number of samples for each modality.

The overall loss function of our proposed method is:
\begin{equation}\label{equ::diff_total}
 \mathcal{L}_{T} = \mathcal{L}_{PBCE} + \lambda \mathcal{L}_{ID}
\end{equation}
where $\lambda$ is a trade-off parameter to balance the losses.
\subsection{Inference Phase}
For inference phase, we can take two evaluation modes which are named as full mode and simplified mode respectively. For the full mode, $D_{RI}$ is used to present the similarity of two images. The smaller $D_{RI}$ is, the more similar two images are.
For the simplified mode, the cosine similarities are calculated between the query and gallery images by using the global features $G_{R}$ and $G_{I}$.
For $P$ query images and $G$ gallery images, the simplified mode need to evaluate the DSCSN for $P+G$ times,
while the full mode need to evaluate additional $P\times G$ Contrastive Correlation Networks. Experimental results in Section~\ref{sec_Exp} show that the full mode can get better performance while the simplified mode is quite efficient and suitable for large scale real applications.

Besides the simplified mode, we think that there are probably  other three types of approximation or simplification method to be considered in the future work.
1) Learning compact binary deep features, which could greatly accelerate the convolution and correlation operations.
2) Designing or searching lightweight architecture for fast inference, like Mobile-Net, Shuffle-Net series.
3) Implementing the convolution/correlation operation by using Fast Fourier Transform (FFT). Existing deep learning libs showed that FFT can accelerate the convolution/correlation process especially when the feature maps has a relatively large size.

\section{Experiments}\label{sec_Exp}
In this section, we report the experimental results to verify the efficacy of the proposed method, including comparison with baseline and state-of-the-art methods, discussion and visualization of our method.
\subsection{Datasets and Evaluation Protocols}

We mainly evaluate our proposed method on two publicly available RGB-IR person re-identification datasets: SYSU-MM01~\cite{wu2017rgb} and RegDB~\cite{nguyen2017person}.

    \noindent \textbf{SYSU-MM01}. It is a large-scale dataset including both indoor and outdoor environments, which is collected by six cameras (four RGB and two near-infrared). And the indoor environments are captured by one IR and two RGB cameras, while the outdoor environments are captured by the other three cameras. It contains $491$ person identities with $287,628$ RGB images and $15,792$ IR images in total, and each person is captured by at least two different spectrum cameras.
    We adopt the most challenging single-shot all-search test mode recommended by~\cite{wu2017rgb} and the same evaluation protocol as in~\cite{wu2017rgb,ye2018hierarchical,ye2018visible,dai2018cross,wang2019learning} is used for fair comparison.
    The training set includes $395$ persons, with $22,258$ RGB images and $11,909$ IR images. The testing set contains $96$ persons, with $3,803$ IR images for query and $301$ selected RGB images selected as the gallery set.

    \noindent \textbf{RegDB}. It is a relatively small-scale dataset containing $412$ persons, which is collected by two aligned cameras (one RGB and one far-infrared). Each person in this dataset has $10$ RGB and $10$ IR images respectively, with $4,120$ RGB images and $4,120$ IR images in total. Following the evaluation protocol adopted in~\cite{ye2018hierarchical,ye2018visible,wang2019learning} for fair comparison, the dataset is randomly split into two halves, one for training and the other for testing. For testing, the RGB images are used as the query set while the IR images as the gallery set.

\noindent\textbf{Evaluation metrics.}
We adopt the standard cumulated matching characteristics (CMC) curve and mean average precision (mAP) which are widely used in the evaluation of conventional ReID problems, to indicate the RGB-IR ReID performance. Note that different from conventional ReID, the query set and the gallery set for RGB-IR ReID task are in different modalities.

\subsection{Implementation Details}
Then, we give the implementation details of our experiments from the following three aspects.

    \noindent\textbf{Input preprocessing.} We resize the resolution of all the images to 256$\times$128. To augment the training data, each training image is padded with 10 zero-valued pixels, so that the size of the images is transformed into 276$\times$148. 256$\times$128 images are randomly cropped and horizontally mirrored from the padded holistic images and then fed into the network.

    \noindent\textbf{Batch Sampling.} $N$ random person identities are firstly selected at each iteration. Then, one RGB image and one IR image of the selected identity are randomly sampled from two different modalities to construct the mini-batch. Thus, totally $2\times N$ images are fed into the network for training at each iteration. In this manner, within each mini-batch, we can select $N$ positive pairs and $r \times N$ negative pairs, where r is the ratio between negative pairs and positive pairs. The total pairs in each batch is $M=N+r\times N$. We set $N=32$ and $r=3$ in our experiments.

    \noindent\textbf{Training.} For the backbone architecture of the dual path network, ResNet-50~\cite{He2015Deep} pre-trained on ImageNet is adopted as backbone. Specifically, the parameters are not shared for the input stem, stage 1 and stage 2 of ResNet-50 during the modality-specific feature extraction step in Figure~\ref{fig:arch}, while they share parameters for the stage 3 and stage 4 which are treated as the feature embedding blocks.
    The maximum number of training epochs is set to 60 for both datasets. The stochastic gradient descent (SGD) optimizer is utilized for optimization. The initial learning rate is set to be 0.1 and then decreased by $1/10$ for the last 30 epochs. We set the trade-off parameter $\lambda = 0.1$ throughout the experiments.

\begin{table}
    \centering
    \caption{Comparison with baseline methods on SYSU-MM01 and RegDB datasets. ``-S,-F'' denotes the simplified and full evaluation mode respectively. ``w/o ID'' denotes our method without ID loss.}
            \resizebox{0.5\textwidth}{!}{
          \begin{tabular}{|c|c|cc|cc|}
            \hline
            -  &- & \multicolumn{2}{|c|}{SYSU-MM01} &\multicolumn{2}{|c|}{RegDB} \\
            \hline
            Methods  &Contrastive features  & cmc-1  & mAP & cmc-1 & mAP \\
            \hline

            VL      &$\times$     & $23.2$ & $27.1$ & $24.0$ & $27.3$ \\
            RKL     &$\times$           & $27.2$ & $29.4$ & $32.1$ & $33.4$ \\
            KVL &$\times$ & $26.3$ & $29.3$ &$34.6$ &$35.7$ \\
            IDL &$\times$      & $28.4$ &$31.8$  &$27.2$ &$31.2$ \\
            \hline
            Ours(w/o ID)-S &\checkmark & $33.3$  & $35.9$ & $50.5$ & $51.1$ \\
            Ours(w/o ID)-F &\checkmark &$34.0$ &$36.3$ &$59.0$ &$57.5$ \\
            Ours-S &\checkmark & ${33.4}$  & ${37.2}$ & ${53.1}$ &${53.5}$ \\
            Ours-F &\checkmark &$\mathbf{35.1}$ &$\mathbf{37.4}$ &$\mathbf{60.8}$ &$\mathbf{60.0}$ \\
            \hline
            \end{tabular}}
    \label{tab:cc2baseline}
\end{table}

\begin{table*}[ht!]
    \centering
    \caption{Comparison with state-of-the-art methods on SYSU-MM01 and RegDB datasets. (\%) }
        \resizebox{0.88\textwidth}{!}{
      \begin{tabular}{|c|ccc|c|ccc|c|}
      \hline
         -  &\multicolumn{4}{|c|}{SYSU-MM01}  &\multicolumn{4}{|c|}{RegDB} \\
         \hline
         Methods  & cmc-1 & cmc-10 & cmc-20  & mAP & cmc-1 & cmc-10 & cmc-20  & mAP \\
        \hline
        HOG~\cite{dalal2005histograms}      & $2.8$ & $18.3$ & $31.9$ &$4.2$   &$13.5$ &$33.2$ &$43.7$ &$10.3$\\
        MLBP~\cite{liao2007learning}    & $2.1$  & $16.2$ & $28.3$ & $3.9$  &$2.0$ &$7.3$ &$10.9$ &$6.8$   \\
        LOMO~\cite{liao2015person}    & $1.8$  & $14.1$ &$26.6$ &$3.5$ &$0.9$ &$2.5$ &$4.1$ &$2.3$ \\
        GSM~\cite{lin2016cross} &$5.3$ &$33.7$ &$53.0$ &$8.0$ &$17.3$ &$34.5$ &$45.3$ &$15.1$ \\
        One-stream Network~\cite{wu2017rgb} &$12.0$ &$49.7$ &$66.7$ &$13.7$ &$13.1$ &$33.0$ &$42.5$ &$14.0$ \\
        Two-stream Network~\cite{wu2017rgb} &$11.7$ &$48.0$ &$65.5$ &$12.9$ &$12.4$ &$30.4$ &$41.0$ &$13.4$ \\
        Zero-padding~\cite{wu2017rgb} &$14.8$ &$54.1$ &$71.3$ &$16.0$ &$17.8$ &$34.2$ &$44.4$ &$18.9$ \\
        TONE~\cite{ye2018hierarchical} &$12.5$ &$50.7$ &$68.6$ &$14.4$ &$16.9$ &$34.0$ &$44.1$ &$14.9$ \\
        HCML~\cite{ye2018hierarchical} &$14.3$ &$53.2$ &$69.2$ &$16.2$ &$24.4$ &$47.5$ &$56.8$ &$20.8$ \\
        BDTR~\cite{ye2018visible} &$17.0$ &$55.4$ &$72.0$ &$19.7$ &$33.5$ &$58.4$ &$67.5$ &$31.8$ \\
        cmGAN~\cite{dai2018cross} &$27.0$ &$67.5$ &$80.6$ &$27.8$ &- &- &- &- \\
        HSME~\cite{hao2019hsme} &$18.0$ &$58.3$ &$74.4$ &$20.0$ &$41.3$ &$65.2$ &$75.1$ &$38.8$ \\
        D-HSME~\cite{hao2019hsme} &$20.7$ &$62.7$ &$78.0$ &$23.1$ &$50.9$ &$73.4$ &$81.7$ &$47.0$ \\
        eBDTR~\cite{ye2019bi} &$27.8$ &$67.3$ &$81.3$ &$28.4$ &$31.8$ &$56.1$ &$66.8$ &$33.2$ \\
        D$^{2}$RL~\cite{wang2019learning} &$28.9$ &$70.6$ &$82.4$ &$29.2$ &$43.4$ &$66.1$ &$76.3$ &$44.1$ \\

        MAC~\cite{ye2019modality} &$33.3$ &$79.0$ &$90.1$ &$36.2$ &$36.4$ &$62.4$ &$71.6$ &$37.0$ \\
        \hline
        Ours(w/o ID)-S &$33.3$ &$76.3$ &$89.0$ &$35.9$ &$50.5$ &$68.3$ &$78.1$ &$51.1$\\
        Ours(w/o ID)-F &$34.0$ &$76.7$ &$87.5$ &$36.3$ &$59.0$ &$75.2$ & $83.5$ &$57.5$ \\
        Ours-S &33.4 &$78.6$ &$89.4$ &$37.2$ &$53.1$ &$72.3$ &$80.2$ &$53.5$ \\
        Ours-F &$\bf{35.1}$ &$77.6$ &$88.9$ &$\bf{37.4}$ &$\bf{60.8}$ &$78.6$ &$85.9$ &$\bf{60.0}$\\
        \hline
        \end{tabular}}

    \label{tab:soa}
\end{table*}

\subsection{Comparison with Baseline Methods}
We compare our approach with four baseline methods on both SYSU-MM01 and RegDB datasets in Table~\ref{tab:cc2baseline}. Note that the same neural architecture is adopted for fair comparison.
Verification loss (VL) ~\cite{zheng2018discriminatively} takes a pair of feature vectors extracted for person images and determine whether they belong to the same person or not.
Ranking loss (RKL) denotes the method proposed in~\cite{ye2018visible}. It is a kind of metric learning method which includes inter-modality and intra-modality top-ranking constraints.
Note that VL and RKL are enforced on top of the global feature $G_{R}$ and $G_{I}$.
We also extend the verification loss to the contrastive kernels $K_{RI}$, denoted as kernel verification loss (KVL).
In addition, ID loss (IDL) without Contrastive Correlation Network is also adopted as the baseline method.

``Ours(w/o ID)-S'' in Table~\ref{tab:cc2baseline} denotes the proposed method without ID loss for using the simplified evaluation mode, while ``-F'' represents using full mode. As can be seen from Table~\ref{tab:cc2baseline} that
1) VL, RKL, KVL and IDL are four variants without contrastive features. It can be clearly seen that with the help of contrastive features, our methods can greatly outperforms those without contrastive features.
2) The proposed method can achieve better performance when ID loss is combined with the pairwise BCE loss in Eqn.~\eqref{equ::diff_total}. 3) Compared with the simplified evaluation mode, the full mode during the inference phase can get better ReID performance for both cmc-1 and mAP.
\begin{table}
    \caption{
    Effect of feature embedding on SYSU-MM01 dataset.
    }
    \centering
        \resizebox{0.45\textwidth}{!}{
            \begin{tabular}{|c|cc|cc|}
                \hline
                - & \multicolumn{2}{|c|}{Ours(w/o ID)} &\multicolumn{2}{|c|}{ID loss} \\
                \hline
                Shared layers     & cmc-1     & mAP  & cmc-1  &mAP \\

                \hline
                None  & $25.9$  & $29.3$  & $20.6$  & $24.8$ \\
                Stage 4             & $29.3$ &  $33.6$  & $27.9$ &  $31.1$\\
                Stage 3-4                   & $\mathbf{33.3}$    & $\mathbf{35.9}$  &$\mathbf{28.4}$    & $\mathbf{31.8}$ \\
                Stage 2-4    & $32.1$ & $34.8$  &$27.6$ & $30.8$ \\
                Stage 1-4  &$32.9$  & $35.6$  &$27.1$  & $30.0$\\
                All   &$32.7$ &$34.7$ &$26.3$ &$30.1$\\
                \hline
                Dual-path~\cite{ye2018visible} &- & - & $25.1$ &  $27.4$ \\
                \hline
            \end{tabular}}

    \label{tab:FME}
\end{table}

\begin{figure*}[ht]
	\centering
		\includegraphics[width=0.8\linewidth]{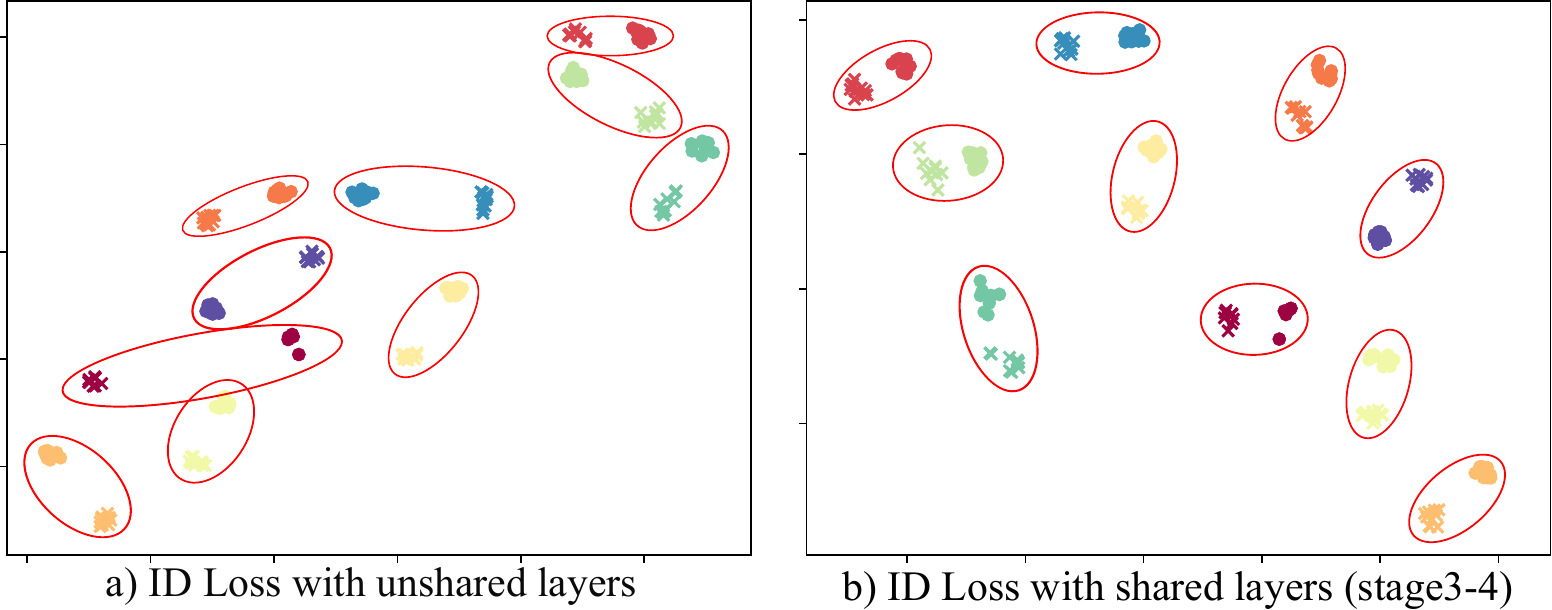}
	\caption{Visualization of the features in common space by using t-SNE~\cite{maaten2008visualizing}. (a) shows the common features obtained with unshared weights embedding function. While (b) shows the case where stage3 and stage4 are shared for feature embedding function. Samples with the same color denote that they are from the same person. The makers "dot" and "cross" denote image from the visible and infrared domain respectively.}
	\label{fig:embed}
\end{figure*}

\begin{figure}[t]
	\centering
		\includegraphics[width=0.47\textwidth]{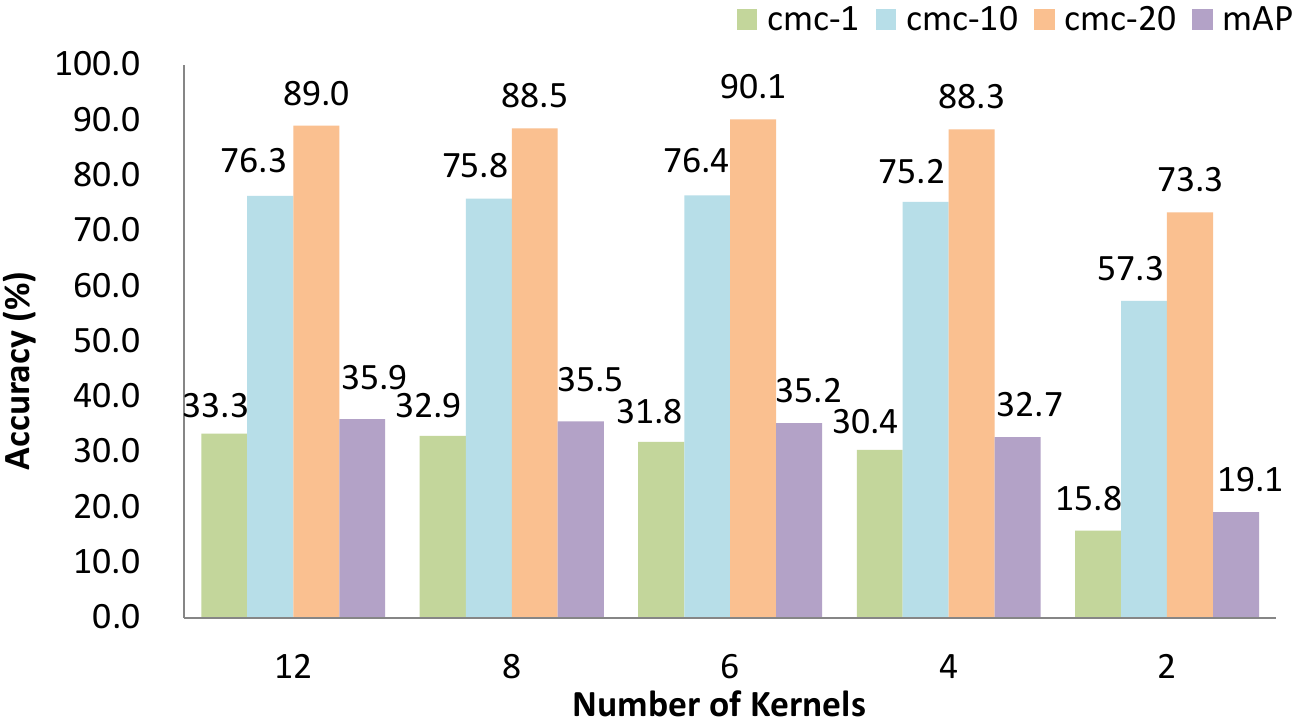}
	\caption{
    Influence of the number of kernels on SYSU-MM01 dataset.
    }
	\label{fig:kerneleffect}
\end{figure}
\begin{figure*}[htb]
	\centering
		\includegraphics[width=0.8\textwidth]{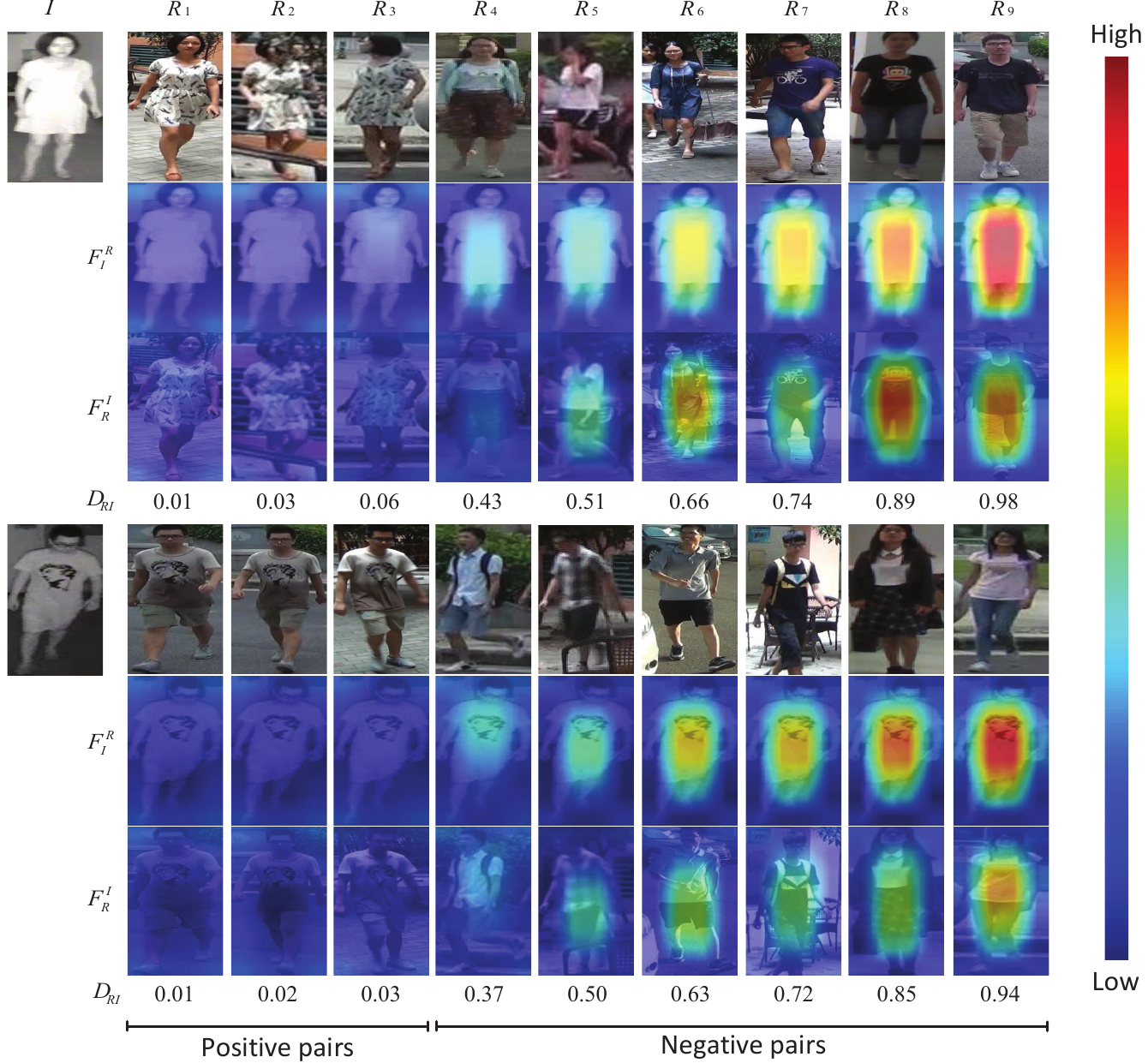}
	\caption{
    Visualization of contrastive feature maps extracted by our proposed method (Ours) on the test set of SYSU-MM01 dataset.}
	\label{fig:feature_map}
\end{figure*}
\begin{figure*}[h!]
	\centering
		\includegraphics[width=0.8\textwidth]{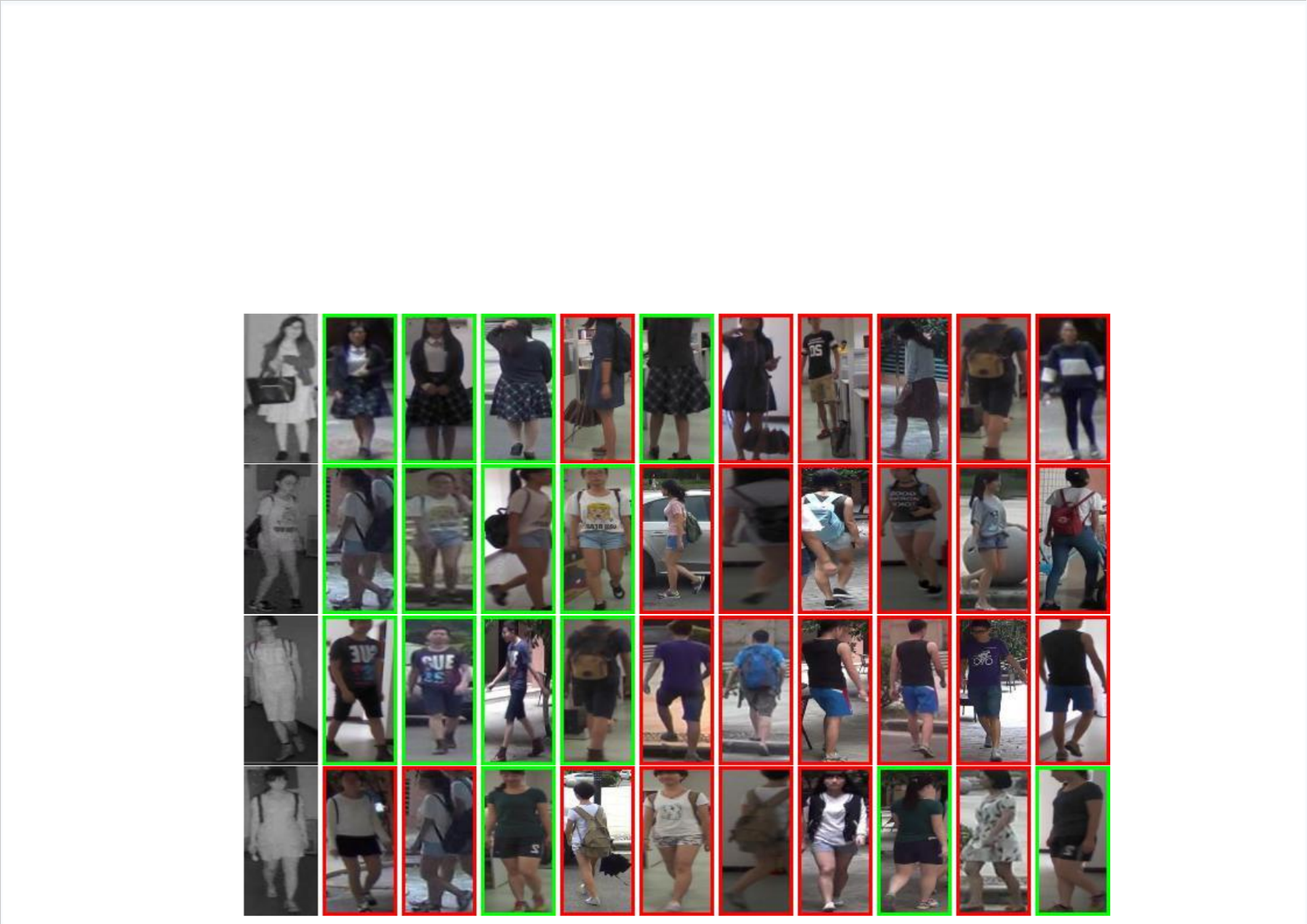}
	\caption{
    Samples of retrieval results on the test set of SYSU-MM01 dataset of our proposed method (Ours-S).}
	\label{fig:result}
\end{figure*}

\subsection{Comparison with State-of-the-art Methods}
We compare our methods with most of the RGB-IR person ReID methods which have reported their results on the SYSU-MM01 and RegDB datasets in Table~\ref{tab:soa}.
Current state-of-the-art methods include: cmGAN~\cite{dai2018cross} proposed to learn modality invariant features via an adversarial way.
HSME~\cite{hao2019hsme} proposed an end-to-end dual-stream hypersphere manifold embedding network with both classification and identification constraint.
D-HSME~\cite{hao2019hsme} modified the weight matrix of Sphere Softmax via Singular Vector Decomposition(SVD).
eBDTR~\cite{ye2019bi} presented a bidirectional center-constrained top-ranking loss to learn discriminative features,
and D$^{2}$RL~\cite{wang2019learning} incorporated a generative model to transform the input into a unified space firstly, then extracted discriminative features from the unified space.
MAC~\cite{ye2019modality} proposed a novel modality-aware collaborative learning method on top of a two-stream network, handling the modality-discrepancy in both feature level and classifier level.
We can see from Table~\ref{tab:soa} that our method significantly outperforms D$^{2}$RL by \textit{17.4\%} cmc-1 and \textit{15.9\%} mAP on RegDB dataset, and outperforms MAC by \textit{1.8\%} cmc-1 and \textit{1.2\%} mAP on SYSU-MM01 dataset.

We compare two types of evaluation modes during inference phase.
As we can see from Table~\ref{tab:soa}, the performance of using full mode is better than using simplified mode. On SYSU-MM01 dataset, the full mode outperforms the simplified mode by $1.7\%$ cmc-1 and $0.2\%$ mAP, while on RegDB dataset, using full mode gets an additional $7.7\%$ cmc-1 and $6.5\%$ mAP improvement. It might because RegDB is a relatively small dataset, and the global features probably can not generalize as well as in a relatively large SYSU-MM01 dataset.

On the other hand, the full mode needs to take more computing resources and is more time-consuming.
Our experiments are based on single NVIDIA RTX-2080Ti GPU. Taking RegDB dateset as an example, it contains $2060$ query images and $2060$ gallery images for testing. First, the DSCSN takes about $24.8$s to extract all query and gallery image features. Then, for simplified mode, the cosine similarity calculations of all query images and gallery images take only about $0.5$s, while for full mode, the Contrastive Correlation Network takes about $5280$s to calculate the difference scores of all query and gallery images ($2060\times 2060$ evaluations of CCN).
So the full mode is less efficient especially when searching in a large gallery set.
\subsection{Discussion}
\noindent\textbf{Effect of feature embedding function.} For RGB-IR person ReID, previous feature embedding methods~\cite{ye2018visible,ye2018hierarchical} use a FC layer to project the modality-specific feature vectors into the common feature space.
Compared with them, we design the feature embedding function as convolution architectures. Table~\ref{tab:FME} shows how the feature embedding function affects the RGB-IR ReID performance.
Note that ResNet-50 is taken as the mapping model which transforms the input into the common feature space, specifically the parameter sharing parts are viewed as the feature embedding function.
It can be clearly seen from Table~\ref{tab:FME} that when stage 3 and stage 4 of ResNet-50 are designed as feature embedding function, it achieves best results no matter what the successive loss function or method (the ID loss or Ours(w/o ID)-S) is used.
Meanwhile, compared with dual-path network~\cite{ye2018visible}, our proposed feature embedding function boosts the performance by $3.3\%$ cmc-1 and $4.4\%$ mAP while ID loss is enforced.

We visualize and compare the 3D feature tensors of shared weights(stage3-4) and unshared weights in common space by using t-SNE~\cite{maaten2008visualizing} in Figure~\ref{fig:embed} (a) and (b).
Figure~\ref{fig:embed} (a) shows the case where feature embedding function are not sharing parameters while in Figure~\ref{fig:embed} (b) sharing stage3 and stage4 is devised as feature embedding function.
A total of 10 persons are randomly selected from the testing set of regDB. Samples with the same color indicate they are the same person. The markers "dot" and "cross" denote image from the visible and infrared domain respectively.
We can observe that single-modality intra-person samples get closer to each other in both Figure~\ref{fig:embed}(a) and Figure~\ref{fig:embed}(b). Comparing with Figure~\ref{fig:embed}(a), Cross-modality intra-person samples move closer in  Figure~\ref{fig:embed}(b) with the help of sharing-weights embedding function.

The initial motivation of designing the feature embedding function as an convolutional architecture is to meet the need of a 3D common space to support the proposed Contrastive Correlation Network (CCN). Thus, how to design or optimize the architecture of both the modality-specific feature extraction and modality-shared feature embedding functions is very important. Traditionally, few research works have been devoted to this topic. In this paper, we explore the problem of architecture design of feature embedding function based on a fixed backbone network (ResNet-50). Although the experimental results showed the potential value of this problem, our experiments are still conducted on a relatively coarse granularity.
We believe that it is a valuable research topic and in the future, we will try to solve the problem with the help of Neural Architecture Search (NAS).
\begin{table}
    \centering
    \caption{Comparison of kernel generator module with our kernel sampling scheme for cross-modality person reID.}
        \resizebox{0.47\textwidth}{!}{
      \begin{tabular}{|c|c|cc|cc|}
        \hline
        -   &- & \multicolumn{2}{|c|}{SYSU-MM01} &\multicolumn{2}{|c|}{RegDB} \\
        \hline
        Methods &Parameters    & cmc-1  & mAP & cmc-1 & mAP \\
        \hline
        KG-S   &$49$M         & $28.9$ & $32.0$ & $28.2$ & $31.5$ \\
        Ours-S   & $25$M         & $33.4$  & $37.2$   & $53.1$  & $53.5$ \\
        Gains    & $-24$M          & $+4.5$   & $+5.2$    & $+24.9$  & $+22.0$ \\
        \hline
        \hline
        KG-F  & $49$M          & $31.9$ & $33.9$ &$34.1$ &$37.4$ \\
        Ours-F  &  $25$M   & $35.1$  & $37.4$  & $60.8$  & $60.0$ \\
        Gains   & $-24$M   &$+3.2$    & $+3.5$   & $+26.7$  & $+22.6$ \\
        \hline
        \end{tabular}}

    \label{tab:kgm}
\end{table}

\begin{table}[htbp]
\centering
\caption{Comparison of our kernel sampling scheme (Ours) and kernel generator module (KG) in~\cite{han2018face} on the face verification task. KG(r) denotes the result reported in~\cite{han2018face}, while KG(*) represents the result of our implementation based on the open source code of~\cite{han2018face}. The performance gap is mainly due to the preprocessing of the dataset. Note that KG(*) and Ours are based on the same pre-processed data and the only difference between them is the kernel generator.
}
\label{tab:face}
    \resizebox{0.48\textwidth}{!} {
     \begin{tabular}{|c|c|c|c|c|}
    \hline
    Methods & Depth & Data &Parameters & mACC on LFW \\
    \hline
    KG(r) & $4$ & WebFace & $64.6$M  & $98.20$ \\
    \hline
    KG(*)    & $4$ & WebFace & $64.6$M & $95.48$ \\
    Ours  & $4$ & WebFace  &$1.6$M & $95.50$ \\
    \hline
    \end{tabular}
    }
\end{table}
\noindent\textbf{Kernel sampling scheme v.s. Kernel generator in~\cite{han2018face}.}
We compare our kernel sampling scheme with kernel generator for cross-modality person reID and face verification in Table~\ref{tab:kgm} and Table~\ref{tab:face} correspondingly.

In Table~\ref{tab:kgm},
KG represents the method which is same as our method, except its kernel generation mechanism changes from kernel sampling scheme to kernel generator.
From Table~\ref{tab:kgm}, we empirically find our proposed kernel sampling scheme can get converged more easily and achieve better accuracy than kernel generator in our task. Compared with the original kernel generator, our kernel generating method boosts the mAP by 3.5\% and cmc-1 by 3.2\% on SYSU-MM01 dataset, and boosts the mAP by 22.6\% and cmc-1 by 26.7\% on RegDB dataset.
It is probably because the original kernel generator in~\cite{han2018face} bring 24M (about 100\%) more parameters than our method and it is easy to over fit on cross-modality person reID datasets.

Besides, to further substantiate the effectiveness and efficiency of our proposed kernel generator module, we replace the original kernel generator in~\cite{han2018face} with our parameter-free kernel sampling scheme. Other settings are kept the same as~\cite{han2018face}. CASIA-WebFace dataset is adopted as the training dataset and mean accuracy (mACC) is calculated on LFW dataset. The results are shown in Table~\ref{tab:face}. Compared with~\cite{han2018face}, our parameter-free kernel sampling scheme gets similar result but has much fewer parameters on the face verification task.

\noindent\textbf{Influence of kernel numbers. }
The size of kernel is set to $3\times 3$ in our experiments,
and the number of kernels generated by Algorithm~\ref{alg:sampling} can vary by setting different $stride\_v$ and $stride\_h$. Here, we investigate how the number of kernels affect the final performance.
The size of the extracted 3D feature tensor $F_{R}$ and $F_{I}$ are $8\times 4 \times 2048$, with $h_{F}=8$, $w_{F}=4$.
$stride\_h$ is fixed to 1, while $stride\_v$ varies in the range of $1$ to $5$, to change the number of kernels from $12$, $8$, $6$, $4$ to $2$.
The results in Figure~\ref{fig:kerneleffect} show that typically more kernels can obtain better performance.

\noindent\textbf{Sharing the FC layer or not in CCN. }
We compare whether sharing the FC layer or not in CCN. The results are shown in Table~\ref{tab:sharing_FC}.
Unshared FC(S) means unsharing the FC layer in CCN and testing using simplify mode,
while Unshared FC(F) expresses testing using full mode.
We can seen that Ours, which sharing the FC layer in CNN, get better performance than Unshared FC.
\begin{table}
    \centering
    \caption{Comparison between unsharing FC and sharing FC in CCN.}
            \resizebox{0.48\textwidth}{!}{
          \begin{tabular}{|c|cc|cc|}
            \hline
            -   & \multicolumn{2}{|c|}{SYSU-MM01} &\multicolumn{2}{|c|}{RegDB} \\
            \hline
            Methods    & cmc-1  & mAP & cmc-1 & mAP \\
            \hline

            Unshared FC (S)          & $31.6$ & $34.9$ & $45.2$ & $44.6$ \\

            Unshared FC (F)          & $33.4$ & $35.4$ & $45.6$ &$46.4$\\

            Ours-S            & $33.4$ & $37.2$ & $53.1$ & $53.5$\\
            Ours-F            & $35.1$ & $37.4$ & $60.8$ & $60.0$\\
            \hline
            \end{tabular}}
    \label{tab:sharing_FC}
\end{table}

\noindent\textbf{Sensitivity of the hyper-parameter $\lambda$.}
We report the experimental results with varying $\lambda$ in Table~\ref{tab:lambda}. It can be seen that when $\lambda=0.1$ the method achieved the best performance and the performances slightly dropped when $\lambda$ fluctuated.
\begin{table}[h]
    \caption{
    Effect of $\lambda$ on SYSU-MM01 and RegDB Dataset.
    }
    \centering
        {
            \begin{tabular}{|c|cc|cc|}
                \hline
                - & \multicolumn{2}{|c|}{SYSU-MM01}
             &\multicolumn{2}{|c|}{RegDB} \\
                \hline
                $\lambda$    & cmc-1     & mAP  & cmc-1  &mAP \\
                \hline
                0.05 & $33.0$ & $36.4$ &$51.7$ & $52.4$\\
                0.1 & $\mathbf{33.4}$ & $\mathbf{37.2}$  & $\mathbf{53.1}$ & $\mathbf{53.5}$\\
                0.2  & $33.4$  & $36.5$ &$52.4$ &$52.9$\\
                0.4   &$32.3$ &$34.8$ &$51.1$ &$51.9$\\
                1.0  & $31.6$  & $34.3$ &$49.7$ &$50.2$\\
                \hline
            \end{tabular}}
    \label{tab:lambda}
\end{table}

\subsection{Visualization}
To qualitatively analyze the proposed method,
we visualize some contrastive feature maps on the test set of SYSU-MM01 as shown in Figure~\ref{fig:feature_map}.
Specifically, an IR query image $I$ is compared to 3 positive and 6 negative RGB person images.

We can observe that the high response of feature maps lies especially on different body,
and the more different they are, the higher response feature maps exhibit.
Comparing $I$ and $R_{3}$, as the two images’ body shape is similar, the contrastive feature maps’ response is relatively low;
Comparing $I$ and $R_{7}$, genders and body forms are quite different, so the contrastive feature maps’ response is higher.

In addition, we also show some ranking results in Figure~\ref{fig:result}. We select some IR images as queries to search from the gallery RGB images of the test set of SYSU-MM01. The images in the first column are query images. The retrieval results are sorted from left to right according to the similarity scores based on the simplified mode.
The last row shows one failure case, where the person in query image carries a bag, while in gallery images the same person does not carry a bag anymore and the top-2 matches in the ranking list are indeed very similar to the query person.

\section{Conclusion}\label{sec_Con}
In this paper, we propose an RGB-IR cross modality person re-identification framework which consists of a Dual-path Spatial-structure-preserving Common Space Network which projects the input images into a 3D tensor common space, and a Contrastive Correlation Network which makes the feature attend to the difference of the paired inputs.
The framework can be end-to-end trained with joint pairwise BCE loss and ID loss. Extensive experiments on two public available benchmark datasets show that our proposed method can outperform state-of-the-art methods by large margins for both full and simplified evaluation modes.

In the future, for better performance we would like to explore the query-guided features which try to match the similar features  between the query and the gallery images besides the dis-similarities. Furthermore, finer-grain semantic parts alignment guided by the explicit or implicit pose estimation modules can be investigated to boost the ReID accuracy.
Additionally, designing lightweight architectures and learning compact features can be studied for extending the proposed method into mobile applications.

\section*{Appendix}
In Section~\ref{sec_Exp}, we mainly focus on the closed-setting \textbf{cross modality} re-identification problem, i.e. when a query image is in RGB/IR modality, the gallery set is constructed with IR/RGB images only.
While in this section, we test our method for the open-set common situation where we do not know whether the gallery set includes the cross modality target person sharing the same ID with the query person or not.
Therefore, the corresponding method should take both cross modality matching and single modality matching into consideration at the same time.

To meet this end, our DSCSN can be trained by additionally inputing RGB-RGB, IR-IR paired training samples for the open-set common situation.
The RGB branch takes the RGB inputs and the IR branch takes the IR inputs.
For the RGB-RGB and IR-IR inputs, the kernel generator of CCN can be calculated with the corresponding obtained feature map pairs.

\begin{figure*}[ht]
	\centering
\includegraphics[width=0.8\linewidth]{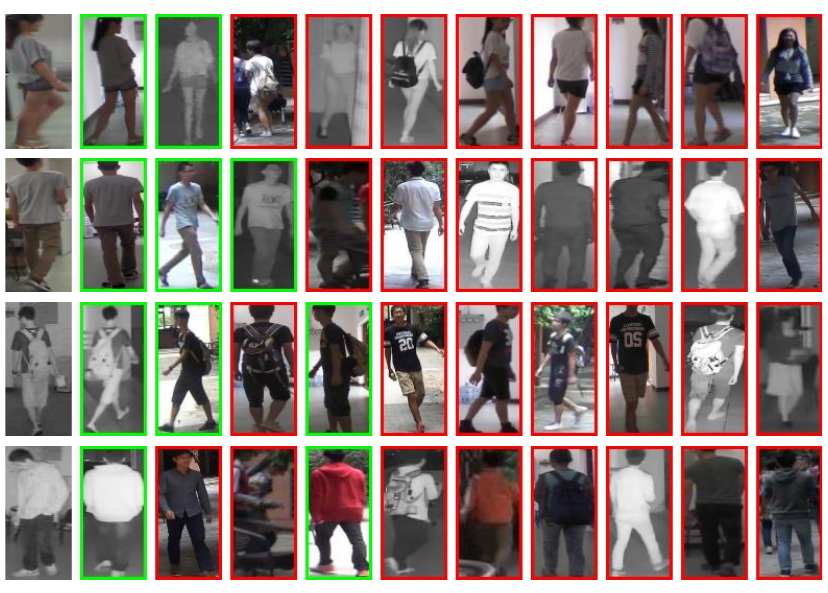}
	\caption{Ranking Results on the Test Set of SYSU-MM01 Dataset for Open-set Protocols. The top 2 rows show the results with RGB queries and the bottom 2 rows show the results with IR queries.}
\label{fig:sample_openset}
\end{figure*}

On SYSU-MM01 Dataset, to show the effectiveness of our proposed method in the open-set common situation, we train our method with $1: 1: 1$ RGB-RGB, RGB-IR and IR-IR paired training samples and test our method with a query set containing $3292$ RGB images from cam1 and cam4, and $1883$ IR images from cam3, and a gallery set containing $3483$ RGB images from cam2 and cam5, and $1920$ IR images from cam6.
Both qualitative and quantitative experimental results are showed as follows,
Table~\ref{tab:open-set}  shows the mAP and cmc ranking results. Figure~\ref{fig:sample_openset} shows several cases of the ranking results including both the RGB and IR images.

\begin{table} [h]
    \caption{
    Experimental results on SYSU-MM01 Dataset for open-set Protocols .
    }
    \centering
            \begin{tabular}{|ccccc|}
                \hline
                cmc-1     & cmc-5  & cmc-10 &cmc-20  &mAP \\
                \hline
                $78.2$ & $94.8$  & $97.7$ & $99.2$ & $65.3$\\
                \hline
            \end{tabular}
    \label{tab:open-set}
\end{table}

\bibliographystyle{IEEEtran}
\bibliography{bibfile1}

\begin{thebibliography}{10}
\providecommand{\url}[1]{#1}
\csname url@samestyle\endcsname
\providecommand{\newblock}{\relax}
\providecommand{\bibinfo}[2]{#2}
\providecommand{\BIBentrySTDinterwordspacing}{\spaceskip=0pt\relax}
\providecommand{\BIBentryALTinterwordstretchfactor}{4}
\providecommand{\BIBentryALTinterwordspacing}{\spaceskip=\fontdimen2\font plus
\BIBentryALTinterwordstretchfactor\fontdimen3\font minus
  \fontdimen4\font\relax}
\providecommand{\BIBforeignlanguage}[2]{{%
\expandafter\ifx\csname l@#1\endcsname\relax
\typeout{** WARNING: IEEEtran.bst: No hyphenation pattern has been}%
\typeout{** loaded for the language `#1'. Using the pattern for}%
\typeout{** the default language instead.}%
\else
\language=\csname l@#1\endcsname
\fi
#2}}
\providecommand{\BIBdecl}{\relax}
\BIBdecl

\bibitem{zheng2016person}
L.~Zheng, Y.~Yang, and A.~G. Hauptmann, ``Person re-identification: Past,
  present and future,'' \emph{arXiv preprint arXiv:1610.02984}, 2016.

\bibitem{li2018adversarial}
X.~Li, A.~Wu, and W.-S. Zheng, ``Adversarial open-world person
  re-identification,'' in \emph{Proceedings of the European Conference on
  Computer Vision (ECCV)}, 2018, pp. 280--296.

\bibitem{wei2018person}
L.~Wei, S.~Zhang, W.~Gao, and Q.~Tian, ``Person transfer gan to bridge domain
  gap for person re-identification,'' in \emph{Proceedings of the IEEE
  Conference on Computer Vision and Pattern Recognition}, 2018, pp. 79--88.

\bibitem{zheng2019joint}
Z.~Zheng, X.~Yang, Z.~Yu, L.~Zheng, Y.~Yang, and J.~Kautz, ``Joint
  discriminative and generative learning for person re-identification,'' in
  \emph{Proceedings of the IEEE Conference on Computer Vision and Pattern
  Recognition}, 2019, pp. 2138--2147.

\bibitem{hou2019vrstc}
R.~Hou, B.~Ma, H.~Chang, X.~Gu, S.~Shan, and X.~Chen, ``Vrstc: Occlusion-free
  video person re-identification,'' in \emph{Proceedings of the IEEE Conference
  on Computer Vision and Pattern Recognition}, 2019, pp. 7183--7192.

\bibitem{yu2019unsupervised}
H.-X. Yu, W.-S. Zheng, A.~Wu, X.~Guo, S.~Gong, and J.-H. Lai, ``Unsupervised
  person re-identification by soft multilabel learning,'' in \emph{Proceedings
  of the IEEE Conference on Computer Vision and Pattern Recognition}, 2019, pp.
  2148--2157.

\bibitem{zhao2019attribute}
Y.~Zhao, X.~Shen, Z.~Jin, H.~Lu, and X.-s. Hua, ``Attribute-driven feature
  disentangling and temporal aggregation for video person re-identification,''
  in \emph{Proceedings of the IEEE Conference on Computer Vision and Pattern
  Recognition}, 2019, pp. 4913--4922.

\bibitem{li2018toward}
K.~Li, Z.~Ding, S.~Li, and Y.~Fu, ``Toward resolution-invariant person
  reidentification via projective dictionary learning,'' \emph{IEEE
  transactions on neural networks and learning systems}, vol.~30, no.~6, pp.
  1896--1907, 2018.

\bibitem{yang2018person}
X.~Yang, P.~Zhou, and M.~Wang, ``Person reidentification via structural deep
  metric learning,'' \emph{IEEE Transactions on Neural Networks and Learning
  Systems}, 2018.

\bibitem{zhou2019person}
R.~Zhou, X.~Chang, L.~Shi, Y.-D. Shen, Y.~Yang, and F.~Nie, ``Person
  reidentification via multi-feature fusion with adaptive graph learning,''
  \emph{IEEE transactions on neural networks and learning systems}, 2019.

\bibitem{xu2015distance}
X.~Xu, W.~Li, and D.~Xu, ``Distance metric learning using privileged
  information for face verification and person re-identification,'' \emph{IEEE
  transactions on neural networks and learning systems}, vol.~26, no.~12, pp.
  3150--3162, 2015.

\bibitem{wu2017rgb}
A.~Wu, W.-S. Zheng, H.-X. Yu, S.~Gong, and J.~Lai, ``Rgb-infrared
  cross-modality person re-identification,'' in \emph{Proceedings of the IEEE
  International Conference on Computer Vision}, 2017, pp. 5380--5389.

\bibitem{ye2018hierarchical}
M.~Ye, X.~Lan, J.~Li, and P.~C. Yuen, ``Hierarchical discriminative learning
  for visible thermal person re-identification,'' in \emph{Thirty-Second AAAI
  Conference on Artificial Intelligence}, 2018.

\bibitem{wang2019learning}
Z.~Wang, Z.~Wang, Y.~Zheng, Y.-Y. Chuang, and S.~Satoh, ``Learning to reduce
  dual-level discrepancy for infrared-visible person re-identification,'' in
  \emph{Proceedings of the IEEE Conference on Computer Vision and Pattern
  Recognition}, 2019, pp. 618--626.

\bibitem{dai2018cross}
P.~Dai, R.~Ji, H.~Wang, Q.~Wu, and Y.~Huang, ``Cross-modality person
  re-identification with generative adversarial training.'' in \emph{IJCAI},
  2018, pp. 677--683.

\bibitem{ye2018visible}
M.~Ye, Z.~Wang, X.~Lan, and P.~C. Yuen, ``Visible thermal person
  re-identification via dual-constrained top-ranking.'' in \emph{IJCAI}, 2018,
  pp. 1092--1099.

\bibitem{wang2017vqa}
P.~Wang, Q.~Wu, C.~Shen, and A.~van~den Hengel, ``The vqa-machine: Learning how
  to use existing vision algorithms to answer new questions,'' in
  \emph{Proceedings of the IEEE Conference on Computer Vision and Pattern
  Recognition}, 2017, pp. 1173--1182.

\bibitem{han2018face}
C.~Han, S.~Shan, M.~Kan, S.~Wu, and X.~Chen, ``Face recognition with
  contrastive convolution,'' in \emph{Proceedings of the European Conference on
  Computer Vision (ECCV)}, 2018, pp. 118--134.

\bibitem{nguyen2017person}
D.~Nguyen, H.~Hong, K.~Kim, and K.~Park, ``Person recognition system based on a
  combination of body images from visible light and thermal cameras,''
  \emph{Sensors}, vol.~17, no.~3, p. 605, 2017.

\bibitem{li2017person}
S.~Li, T.~Xiao, H.~Li, B.~Zhou, D.~Yue, and X.~Wang, ``Person search with
  natural language description,'' in \emph{Proceedings of the IEEE Conference
  on Computer Vision and Pattern Recognition}, 2017, pp. 1970--1979.

\bibitem{zheng2017dual}
Z.~Zheng, L.~Zheng, M.~Garrett, Y.~Yang, and Y.-D. Shen, ``Dual-path
  convolutional image-text embedding with instance loss,'' \emph{arXiv preprint
  arXiv:1711.05535}, 2017.

\bibitem{zhou2017attention}
T.~Zhou, M.~Chen, J.~Yu, and D.~Terzopoulos, ``Attention-based natural language
  person retrieval,'' in \emph{Proceedings of the IEEE Conference on Computer
  Vision and Pattern Recognition Workshops}, 2017, pp. 27--34.

\bibitem{ha2016hypernetworks}
D.~Ha, A.~Dai, and Q.~V. Le, ``Hypernetworks,'' \emph{arXiv preprint
  arXiv:1609.09106}, 2016.

\bibitem{kang2017incorporating}
D.~Kang, D.~Dhar, and A.~Chan, ``Incorporating side information by adaptive
  convolution,'' in \emph{Advances in Neural Information Processing Systems},
  2017, pp. 3867--3877.

\bibitem{jia2016dynamic}
X.~Jia, B.~De~Brabandere, T.~Tuytelaars, and L.~V. Gool, ``Dynamic filter
  networks,'' in \emph{Advances in Neural Information Processing Systems},
  2016, pp. 667--675.

\bibitem{li2018high}
B.~Li, J.~Yan, W.~Wu, Z.~Zhu, and X.~Hu, ``High performance visual tracking
  with siamese region proposal network,'' in \emph{Proceedings of the IEEE
  Conference on Computer Vision and Pattern Recognition}, 2018, pp. 8971--8980.

\bibitem{He2015Deep}
K.~He, X.~Zhang, S.~Ren, and J.~Sun, ``Deep residual learning for image
  recognition,'' 2015.

\bibitem{dalal2005histograms}
N.~Dalal and B.~Triggs, ``Histograms of oriented gradients for human
  detection,'' 2005.

\bibitem{liao2007learning}
S.~Liao, X.~Zhu, Z.~Lei, L.~Zhang, and S.~Z. Li, ``Learning multi-scale block
  local binary patterns for face recognition,'' in \emph{International
  Conference on Biometrics}.\hskip 1em plus 0.5em minus 0.4em\relax Springer,
  2007, pp. 828--837.

\bibitem{liao2015person}
S.~Liao, Y.~Hu, X.~Zhu, and S.~Z. Li, ``Person re-identification by local
  maximal occurrence representation and metric learning,'' in \emph{Proceedings
  of the IEEE conference on computer vision and pattern recognition}, 2015, pp.
  2197--2206.

\bibitem{lin2016cross}
L.~Lin, G.~Wang, W.~Zuo, X.~Feng, and L.~Zhang, ``Cross-domain visual matching
  via generalized similarity measure and feature learning,'' \emph{IEEE
  transactions on pattern analysis and machine intelligence}, vol.~39, no.~6,
  pp. 1089--1102, 2016.

\bibitem{hao2019hsme}
Y.~Hao, N.~Wang, J.~Li, and X.~Gao, ``Hsme: hypersphere manifold embedding for
  visible thermal person re-identification,'' in \emph{Proceedings of the AAAI
  Conference on Artificial Intelligence}, vol.~33, 2019, pp. 8385--8392.

\bibitem{ye2019bi}
M.~Ye, X.~Lan, Z.~Wang, and P.~C. Yuen, ``Bi-directional center-constrained
  top-ranking for visible thermal person re-identification,'' \emph{IEEE
  Transactions on Information Forensics and Security}, 2019.

\bibitem{ye2019modality}
M.~Ye, X.~Lan, and Q.~Leng, ``Modality-aware collaborative learning for visible
  thermal person re-identification,'' in \emph{Proceedings of the 27th ACM
  International Conference on Multimedia}, 2019, pp. 347--355.

\bibitem{zheng2018discriminatively}
Z.~Zheng, L.~Zheng, and Y.~Yang, ``A discriminatively learned cnn embedding for
  person reidentification,'' \emph{ACM Transactions on Multimedia Computing,
  Communications, and Applications (TOMM)}, vol.~14, no.~1, p.~13, 2018.

\bibitem{maaten2008visualizing}
L.~v.~d. Maaten and G.~Hinton, ``Visualizing data using t-sne,'' \emph{Journal
  of machine learning research}, vol.~9, no. Nov, pp. 2579--2605, 2008.

\end{thebibliography}

\begin{IEEEbiography}[{\includegraphics[width=1in,height=1.25in,clip,keepaspectratio]{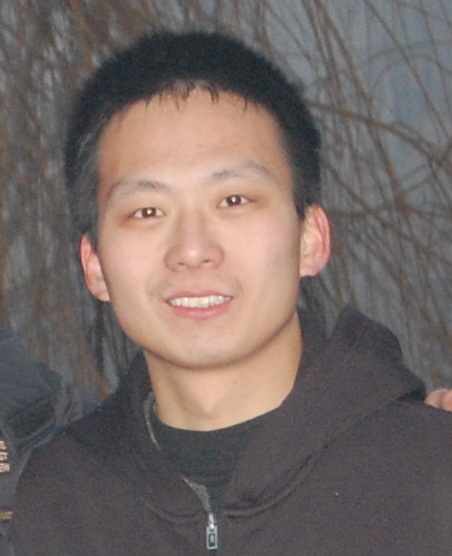}}]{Shizhou Zhang} received a B.E. and Ph.D. degree from Xi'an Jiaotong University, Xi'an, China, in 2010 and 2017, respectively. Currently, he is with Northwestern Polytechnical University as an assistant professor. His research interests include content-based image analysis, pattern recognition and machine learning, specifically in the areas of deep learning based vision tasks such as image classification, object detection, re-identification and semantic parsing.
\end{IEEEbiography}

\begin{IEEEbiography}[{\includegraphics[width=1in,height=1.25in,clip,keepaspectratio]{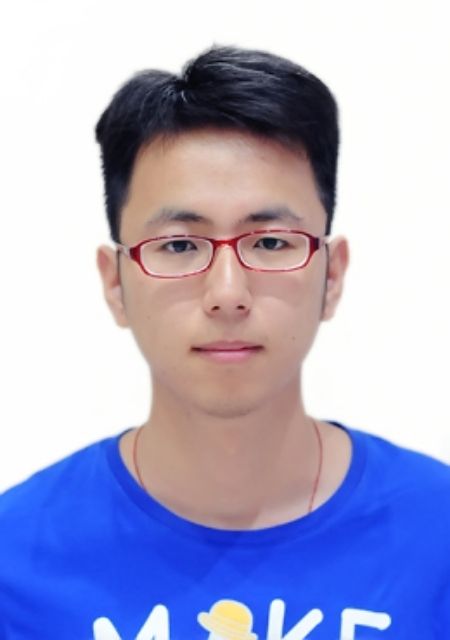}}]
{Yifei Yang} received a B.E. degree from Tianjin University, Tianjin, China, in 2016.
Currently, he is a Master student at Northwestern Polytechnical University, his research direction includes image analysis, pattern recognition and machine learning.
\end{IEEEbiography}

\begin{IEEEbiography}[{\includegraphics[width=1in,height=1.25in,clip,keepaspectratio]{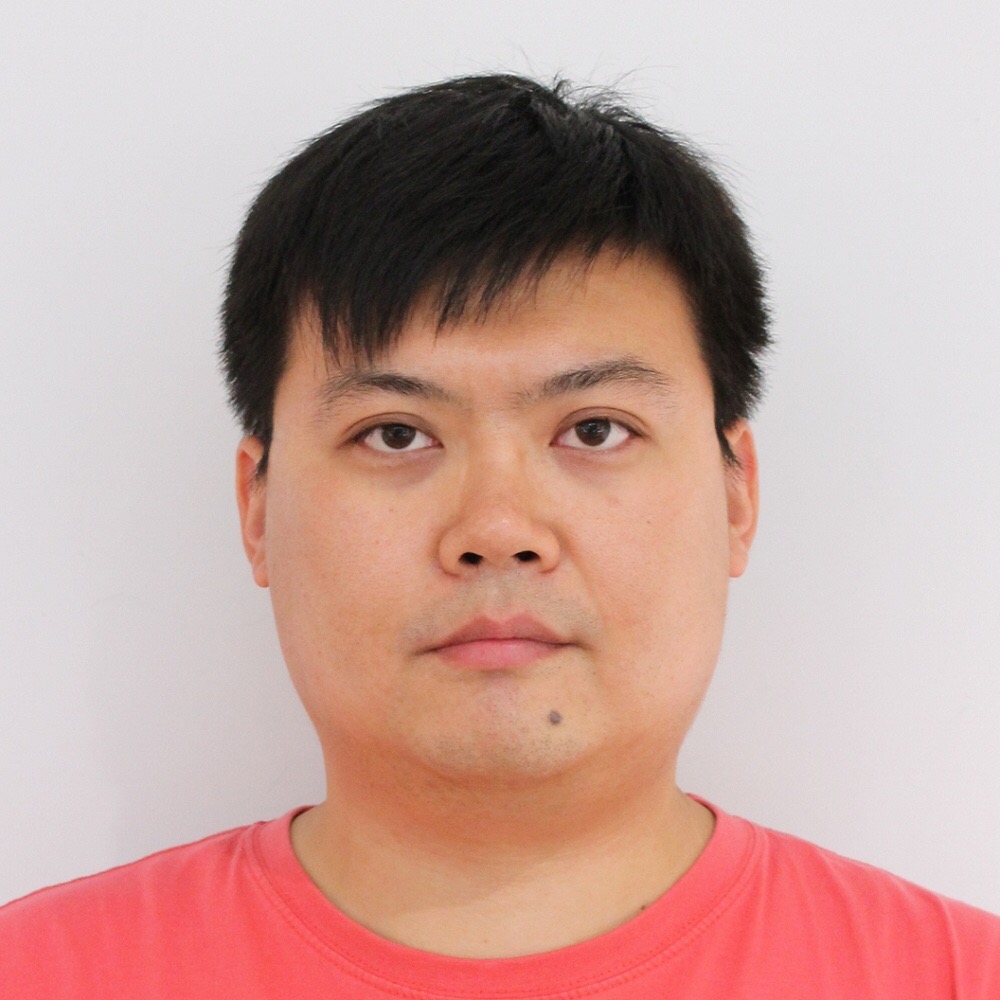}}]
{Peng Wang} is a Professor at School of Computer Science, Northwestern Polytechnical University, China. He was with School of Computer Science, the University of Adelaide for about four years. His research interests are computer vision, machine learning and artificial intelligence. He received a Bachelor in electrical engineering and automation, and a PhD in control science and engineering from Beihang University (China) in 2004 and 2011, respectively.
\end{IEEEbiography}

\begin{IEEEbiography}[{\includegraphics[width=1in,height=1.25in,clip,keepaspectratio]{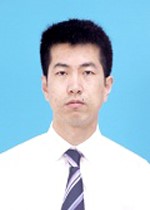}}]
{Guoqiang Liang} received the B.S. in automation and the Ph.D. degrees in pattern recognition and intelligent systems from Xi$'$an Jiaotong University (XJTU), Xi$'$an, China in 2012 and 2018 respectively. From Mar. to Sep. 2017, he was a visiting Ph.D. Student with the University of South Carolina, Columbia, SC, USA. Currently, he is doing the Post-Doctoral Research at the School of Computer Science and Engineering, Northwestern Polytechnical University, Xi$'$an, China. His research interests include human pose estimation and human action classification.
\end{IEEEbiography}

\begin{IEEEbiography}[{\includegraphics[width=1in,height=1.25in,clip,keepaspectratio]{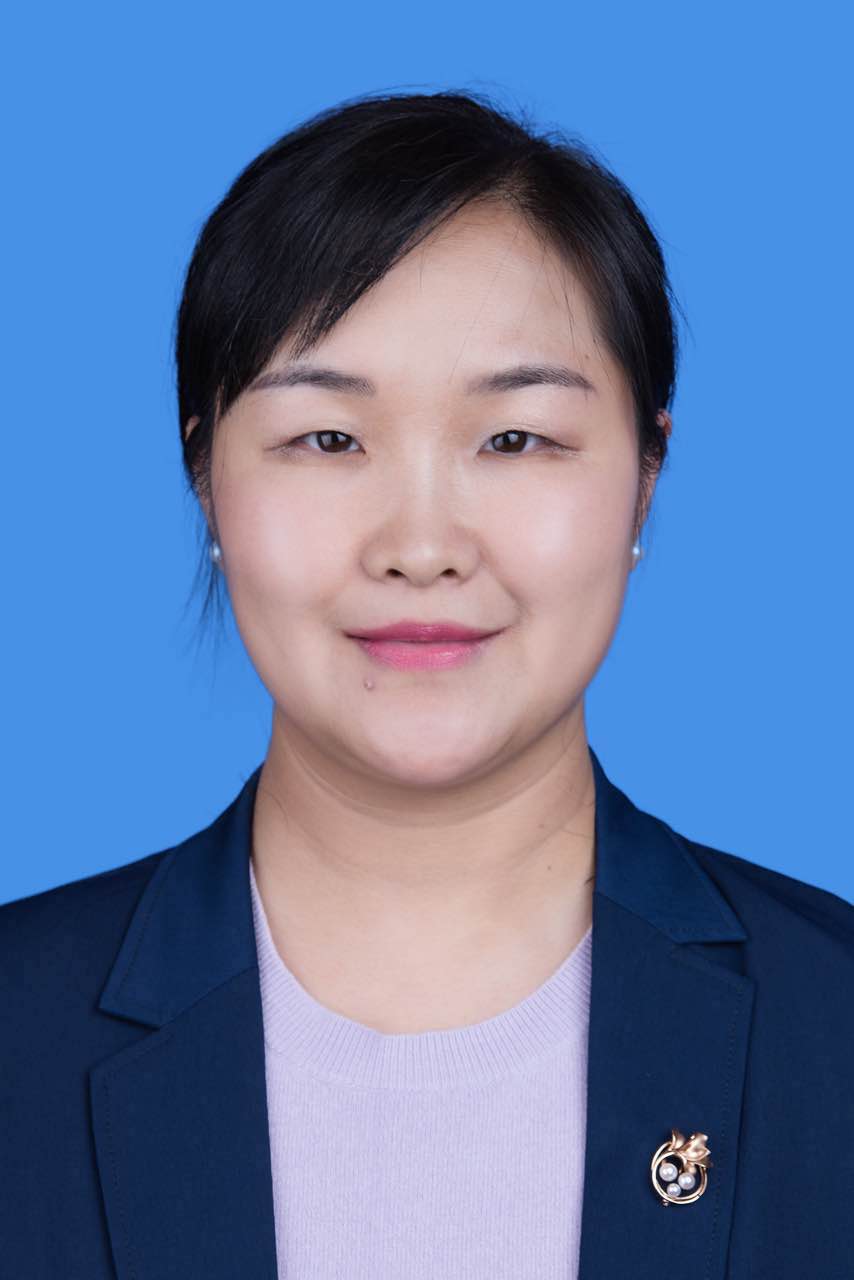}}]
{Xiuwei Zhang}  received the M.S. and Ph.D. degrees in computer science and technology from Northwestern Polytechnical University, Xian, China, in 2007 and 2011, respectively. She is currently an Associate Professor with the Shanxi Provincial Key Laboratory of Speech and Image Information Processing, Northwestern Polytechnical University, Her research interests include multisensor information fusion, image registration,segmentation,and target recognition.
\end{IEEEbiography}

\begin{IEEEbiography}[{\includegraphics[width=1in,height=1.25in,clip,keepaspectratio]{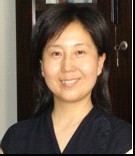}}]
{Yanning Zhang} (SM'10)
received her B.S. Degree from Dalian University of Science and Engineering in 1988, M.S. and Ph.D. degrees from Northwestern Polytechnical University in 1993 and 1996, respectively. She is presently a Professor of School of Computer Science, Northwestern Polytechnical University. She is also the organization chair of the Ninth Asian Conference on Computer Vision (ACCV2009).
Her research work focuses on signal and image processing, computer vision and pattern recognition. She has published over 200 papers in international journals,
conferences and Chinese key journals.
\end{IEEEbiography}

\end{document}